\documentclass[sigconf,letterpaper,nonacm,natbib=false]{acmart}
\AtBeginDocument{%
  }

\usepackage{dblfloatfix}
\usepackage[T1]{fontenc}
\usepackage{xspace}
\usepackage[
  backend=biber,
  datamodel=acmdatamodel,
  style=acmnumeric,
  sorting=none,
  maxbibnames=1,
  minbibnames=1,
  maxcitenames=2,
  mincitenames=1
]{biblatex}

\newcommand{\systemname}{\systemnamens\xspace}
\newcommand{\systemnamens}{\textsf{HiCoTraj}}

\begin{document}

\title[\systemname]{\systemname: Zero-Shot Demographic Reasoning via \\ Hierarchical Chain-of-Thought Prompting from Trajectory }

\author{Junyi Xie, Yuankun Jiao, Jina Kim, Yao-Yi Chiang}
\email{{xie00422, jiao0052, kim01479, yaoyi}@umn.edu}
\affiliation{%
  \institution{University of Minnesota}
  \city{Minneapolis}
  \state{Minnesota}
  \country{USA}
}

\author{Lingyi Zhao, Khurram Shafique}
\email{{lzhao, kshafique}@novateur.ai} 
\affiliation{%
  \institution{Novateur Research Solutions}
  \city{Ashburn}
  \state{Virginia}
  \country{USA}}

\renewcommand{\shortauthors}{Xie et al.}

\begin{abstract}
Inferring demographic attributes such as age, sex, or income level from human mobility patterns enables critical applications such as targeted public health interventions, equitable urban planning, and personalized transportation services. Existing mobility-based demographic inference studies heavily rely on large-scale trajectory data with demographic labels, leading to limited interpretability and poor generalizability across different datasets and user groups. We propose \textbf{\systemname} (Zero-Shot Demographic Reasoning via \textbf{Hi}erarchical \textbf{C}hain-\textbf{o}f-Thought Prompting from \textbf{Traj}ectory), a framework that leverages LLMs' zero-shot learning and semantic understanding capabilities to perform demographic inference without labeled training data. \systemname transforms trajectories into semantically rich, natural language representations by creating detailed activity chronicles and multi-scale visiting summaries. Then \systemname uses a novel hierarchical chain of thought reasoning to systematically guide LLMs through three cognitive stages: factual feature extraction, behavioral pattern analysis, and demographic inference with structured output. This approach addresses the scarcity challenge of labeled demographic data while providing transparent reasoning chains. Experimental evaluation on real-world trajectory data demonstrates that \systemname achieves competitive performance across multiple demographic attributes in zero-shot scenarios.

\end{abstract}




\maketitle

\vspace{-.1in}
\section{Introduction}

Trajectory data capture diverse human mobility aspects and are widely used in tasks such as mobility prediction~\cite{c:liu2024urban,c:wang2022inferring}, enabling key applications: targeted public health interventions, equitable urban planning, personalized transportation services. 
The accuracy of trajectory-inferred demographic and behavioral characteristics directly determines the quality of downstream applications built upon these attributes.
For example, ~\cite{c:stanford2024numosim,c:yue2019detect,c:yue2021vambc,c:lin2024unified} extracted features such as geographic contexts, moving patterns, and activity types from trajectories, which originally consisted only of timestamps and locations, to support subsequent tasks including human mobility clustering and anomaly detection.
Demographic attributes, which capture various socioeconomic characteristics, correlate strongly with mobility trajectories~\cite{c:wu2019inferring}.
Existing studies primarily rely on machine and deep learning (M/DL) techniques to infer demographics from trajectories, but face limited interpretability and often ignore domain knowledge.
Moreover, M/DL methods typically require supervised training on large, high-quality labeled datasets for favorable accuracy; 
Yet, constructing such datasets is particularly challenging in the mobility domain, where trajectories are inherently sparse time-series signals and demographic groups are highly diverse ~\cite{c:jin2023time,c:wu2019inferring}. To the best of our knowledge, prior research has not explored unsupervised demographic inference from mobility data.

\begin{figure*}[!b]
  \vspace{-.1in}
  \includegraphics[width=\textwidth]{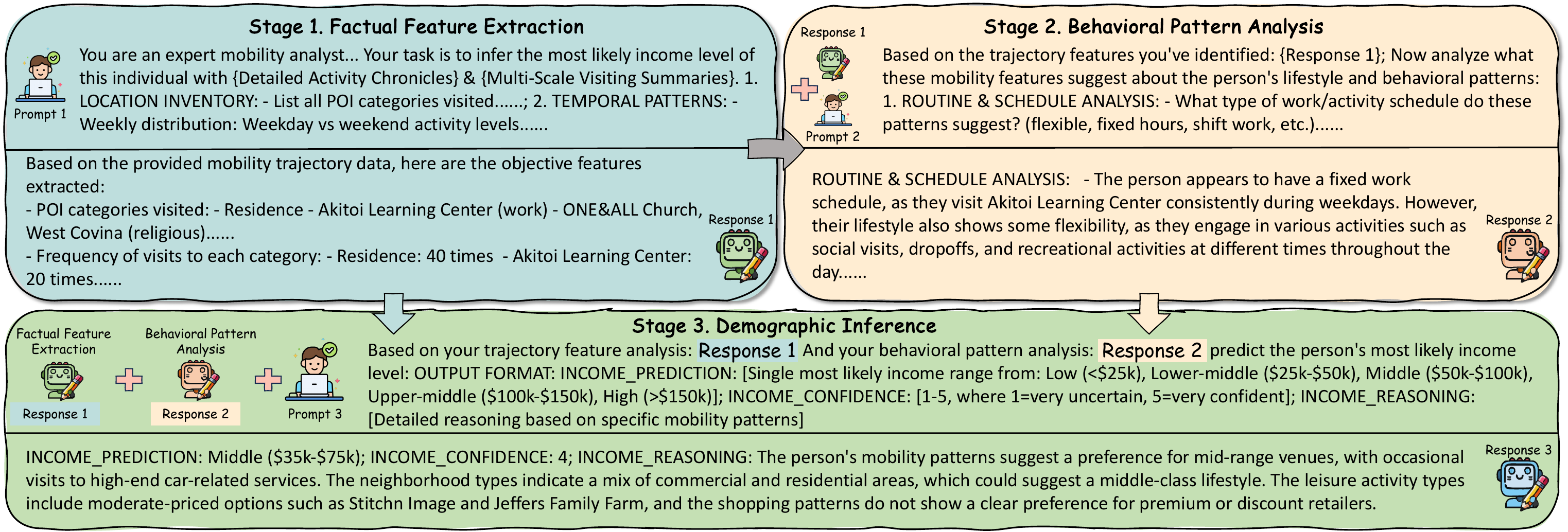}
  \vspace{-.1in} 
  \caption{Hierarchical CoT reasoning procedure in \systemname (Zero-Shot Demographic Reasoning via Hierarchical Chain-of-Thought Prompting from Trajectory).   
  \vspace{-.1in}}
  \label{fig:framework}
\end{figure*}

Trajectory-based demographic inference faces three key challenges—scarcity of labeled datasets, the diversity of demographic groups, and the need for interpretable reasoning. These challenges demand methods that can generalize from limited supervision while capturing semantic and contextual knowledge.
Large Language Models (LLMs) are particularly well-suited to address these challenges. Beyong their transformative impact across multiple domains~\cite{c:xia2024beyond},
LLMs demonstrate the ability to connect language with real-world concepts~\cite{c:wang2025limits} and encode vast domain knowledge usable for reasoning about mobility behaviors.
With Chain-of-Thought (CoT) prompting, they also exhibit improved logic reasoning~\cite{c:sprague2024cot}. Researchers have already started to leverage these strengths for inferring demographic and behavioral characteristics from mobility data~\cite{c:gong2024mobility,c:yu2024chain}, showing that LLMs can provide more interpretable insights into mobility trajectories ~\cite{c:yu2024chain}.
In addition, studies show that LLMs are well suited for zero-shot and few-shot reasoning tasks~\cite{c:brown2020language,c:kojima2022large,c:zhang2025igniting}, aligning directly with the challenge of performing demographic inference in supervised settings where labeled trajectory data are lacking. 

Building on these insights, we introduce \systemname (Zero-Shot Demographic Reasoning via \textbf{Hi}erarchical \textbf{C}hain-\textbf{o}f-Thought Prompting from \textbf{Traj}ectory), a zero-shot demographic reasoning framework that harnesses LLMs’ semantic understanding capabilities via hierarchical CoT prompting. \systemname has two core components: contextual mobility narrative generation, which transforms numerical trajectories into structured weekly activity chronicles, and hierarchical CoT reasoning, which guides LLMs through three stages: factual feature extraction, behavioral pattern analysis, and demographic inference. This training-free approach circumvents the scarcity of labeled demographic data while providing transparent and interpretable inference. 

\vspace{-.1in}
\section{Related Work}

In recent years, the rapid advancement of artificial intelligence has driven a boom in research to infer user demographics from mobile behavioral data, including mobility data~\cite{c:purificato2024user}. Early studies propose diverse dimensions for feature engineering of trajectories and further integrate various M/DL techniques to achieve demographic inference~\cite{c:wang2022inferring,c:gao2024income,c:wu2019inferring}. For example, ~\cite{c:wu2019inferring} proposes a demographic inference framework by mining spatiotemporal and semantic contextual information and feeding it into classical supervised learning algorithms for inference, such as XGBoost and random forests. Most of such studies follow a data-driven paradigm, focusing on feature engineering and representation learning from trajectories. However, this paradigm shows a critical limitation: poor generalizability between user groups such as urban residents and travelers~\cite{c:liu2024urban}. 

Recently, researchers have started to leverage the semantic understanding and contextual reasoning capabilities of language models for behavioral and demographic traits inference. For example, ~\cite{c:gong2024mobility} captures the intention of each check-in point by prioritizing relevant check-in records, and further matches appropriate prompts from a shared pool to enable the LLM to comprehend travel preferences. Complementarily, ~\cite{c:yu2024chain} investigates inferring demographic features such as age, race, and gender using facial image inputs to multimodal models, a method that shows strong advantages in zero-shot learning, interpretability, and handling uncurated "in-the-wild" inputs. ~\cite{c:yu2024chain} also proposes a CoT-augmented prompting approach to address the issue that the language-model-generated answers do not fit the ground truth categories in general classification tasks. Despite these advances, the use of LLMs for demographic inference with trajectory as input remains underexplored, leaving their potential value and technical strengths insufficiently realized. CoT prompting enables models to perform multi-stage reasoning by thinking step by step, achieving advanced performance in tasks such as arithmetic and symbolic reasoning~\cite{c:brown2020language}. 
 
Although a few studies have begun to apply CoT with LLMs in human mobility scenarios, to the best of our knowledge, no existing work exploits the zero-shot or few-shot reasoning strengths of LLMs with CoT for unsupervised inference on trajectory data. Moreover, outside verifiable domains such as mathematics and science, existing CoT frameworks lack robust reasoning capabilities in general scenarios~\cite{c:wang2503multimodal}. Specifically, ~\cite{c:sprague2024cot} shows that the accuracy of reasoning remains nearly the same regardless of whether CoT is used, unless the question contains symbolic operations and reasoning. Therefore, we aim to design CoT frameworks tailored to the characteristics of trajectory data and real-world human mobility scenarios.

\section{\systemname}

As shown in Figure~\ref{fig:framework}, we introduce \systemname, a zero-shot framework that improves interpretability and eliminates the need for labeled data, making it suitable for personalized mobility applications and cross-domain deployment scenarios~\cite{c:al2016user}. Our framework consists of two components: contextual mobility narrative generation and hierarchical CoT reasoning, organized into three stages. 

\paragraph{\textbf{Contextual Mobility Narrative Generation}}
\systemname transforms stay points and POI data into textual prompt formats that preserve the essential granular details, such as visited place names, for zero-shot demographic inference. The input comprises two components: stay point records containing temporal information (start time, end time) and spatial information with corresponding matched POIs; and POI metadata including venue names and pre-assigned activity types. \systemname generates comprehensive weekly activity narratives that maintain venue-specific information and temporal contexts. The weekly activity narratives include: 
(1) Detailed Activity Chronicles. To describe micro-level visiting decisions (specific venue choices, visit durations), \systemname constructs weekly visit documents by concatenating all visiting records of an individual in a week, including precise venue names, date, start-end timestamps, durations, and activity types. For instance: "Monday, January 29 (Weekday) - 09:10-10:14 (63 mins): Bear Wire - Work, Services, DropOff." 
(2) Multi-Scale Visiting Summaries. At the same time, to describe macro-level lifestyle patterns (routine consistency, work-life balance), \systemname generates weekly statistics that analyze types and occurrences by visiting behavior or time. This includes visitation frequency analysis, activity time distribution, and temporal pattern contrasts (e.g., "Average activities on weekdays: 2.7, weekends: 3.5"). \systemname feeds the generated textual narratives into the LLMs, representing micro and macro-level features used for hierarchical CoT reasoning.

\paragraph{\textbf{Hierarchical CoT Reasoning}} \systemname consists of a three-stage hierarchical CoT reasoning framework that systematically decomposes the complex demographic inference task into manageable cognitive stages with increasing levels of abstraction. \systemname establishes a clear cognitive progression: \textbf{Stage 1: Factual Feature Extraction} focuses on factual trajectory analysis without interpretation; \textbf{Stage 2: Behavioral Pattern Analysis} transforms factual observations into lifestyle interpretations and behavioral reasoning to bridge the gap between trajectory features and demographic inference; and \textbf{Stage 3: Demographic Inference} synthesizes the contextual understanding into systematic demographic inference. This enables LLMs to build robust reasoning chains from concrete trajectory observations to abstract demographic conclusions, addressing the explosion of reasoning complexity. Here, we take income as an example to explain the framework.

\noindent\textbf{Stage 1: Factual Feature Extraction} 
This stage limits the model to only descriptive analysis of visiting narratives, without in-depth interpretation or reasoning. Given the detailed daily chronicles and weekly behavioral synthesis, the model systematically extracts: 

The model systematically extracts four types of features: location inventory (POI categories, venue types), temporal patterns (activity hours, weekly distributions, routine consistency), spatial characteristics (geographic distribution), and sequence observations (location transitions, daily/weekly regularities).
The intuition is that the model is not inferring hidden attributes after mobility input, but simply extracting explicitly patterns like timestamps and venue names from narratives. LLMs excel at such text-based pattern recognition and semantic categorization, similar to summarization tasks where hallucination is minimal. By setting explicit boundaries, the framework grounds outputs in observable evidence, ensuring faithful generation.

\noindent\textbf{Stage 2: Behavioral Pattern Analysis} This stage transforms the factual features extracted in Stage 1 into lifestyle interpretations and behavioral reasoning to bridge the gap between trajectory observations and demographic inference. 

The framework analyzes five behavioral dimensions: temporal patterns (work-life structure), economic patterns (spending preferences), social patterns (lifestyle choices), spatial patterns (living environment), and stability patterns (routine consistency).
For instance, combining regular weekday visits to budget venues (economic behavior) with consistent temporal patterns (routine stability) and limited spatial radius (urban characteristics) provides a comprehensive behavioral modeling. This multi-dimensional behavioral abstraction layer ensures a robust contextual foundation for subsequent demographic inference while maintaining clear boundaries between lifestyle interpretation and demographic prediction.

\noindent\textbf{Stage 3: Demographic Inference} This stage uses full responses from Stages 1 and 2 as contextual input, then performs systematic demographic inference through a structured evaluation framework. For income inference, the model evaluates five specific income indicators on a 1-10 scale: location economic levels (luxury, mid-range, budget), neighborhood characteristics (affluent, middle-class, working-class), leisure cost levels types (expensive, moderate, free), shopping patterns (premium, mid-range, discount retailers), and commuting patterns (private transport, rideshare, public transit). The model then synthesizes these evaluations into standardized outputs: a categorical income prediction using six predefined brackets (Very Low < \$15k, Low <\$15k-\$35k, Middle \$35k-\$75k, Upper-middle \$75k-\$125k, High \$125k-\$200k, Very high >\$200k); explicit confidence scoring (1-5 scale); detailed evidence-based reasoning linking specific mobility patterns to conclusions; ranked alternative predictions with supporting rationale. This structured design converts rich contextual understanding from hierarchical reasoning into evaluable predictions, while preserving transparency via explicit evidence chains.

\vspace{-.1in}
\section{Experiments}
\paragraph{\textbf{Experimental Setup}}
We evaluate \systemname on the NUMOSIM~\cite{c:stanford2024numosim} dataset. The dataset provides spatiotemporal information and each agent's demographic labels, including age, sex, income, and education level. Specifically, spatiotemporal information consists of start/end timestamps, POI (longitude and latitude with venue names), and pre-labeled POI activity types for each POI. For the NUMOSIM dataset, they process raw trajectories to identify POIs and their associated activity types. When dealing with other datasets with only raw trajectory (longitude and latitude), we can use trajectory mining methods to obtain rich spatial context as input to \systemname. For our experiments, We randomly sample 6,000 agents using a fixed random seed to ensure reproducible experimental results.
We evaluate Mistral-7B~\cite{c:jiang2023mistral7b} and Qwen3-8B~\cite{c:qwen3} to assess framework generalizability across multiple LLMs. We employ model-specific response parsers to ensure consistency of demographic categories and output formats across models. We assess performance using standard metrics for classification evaluation: Accuracy and F1 score across age, income, and education. 

\vspace{-.1in}

\paragraph{\textbf{Experimental Results and Discussion}}
Table~\ref{tab:main_results} presents the experimental results on zero-shot demographic inference tasks, including age, income, and education. Age prediction (4 categories) demonstrates good performance with accuracy rates of 0.442 and 0.366 for Mistral-7B and Qwen3-8B, respectively. Income prediction (6 categories) achieves better accuracy (0.293 and 0.285) than education prediction (5 categories, 0.254 and 0.238), despite having more classification categories. These results indicate that age- and income-related mobility patterns are more distinct, with income reflected in choices of venues, transport, and residential areas, while educational patterns are harder to detect from trajectory data.

F1 scores highlight significant class-imbalance challenges. The model struggles to balance precision and recall across demographic categories in zero-shot settings. The gap between accuracy and F1 shows that minority classes receive less reliable predictions, lowering F1 scores. \systemname tests the feasibility of LLM-based demographic prediction and provides a transparent, reproducible baseline.

\begin{figure}[h]
  \centering
  \includegraphics[width=\columnwidth]{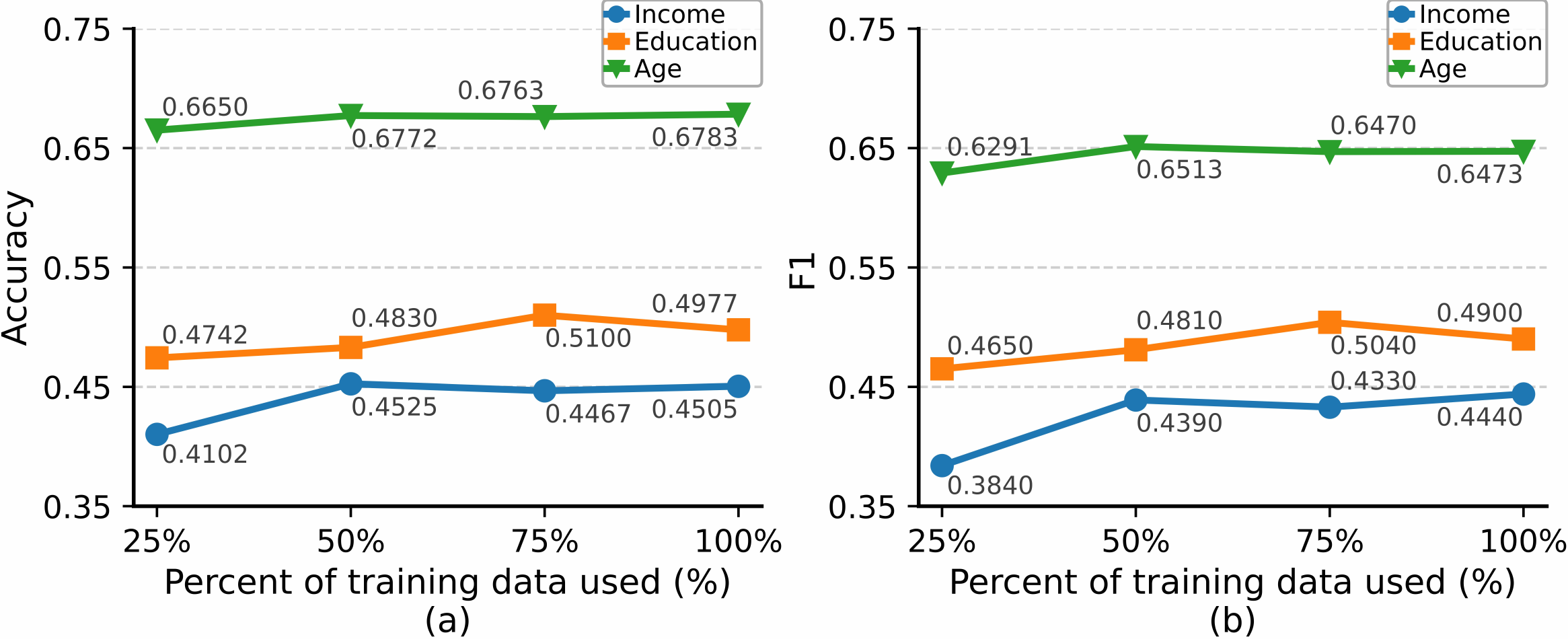}
  \caption{Performance of supervised transformer baseline across varying training data sizes. (a) Accuracy for income, education, and age prediction. (b) F1 score for income, education, and age prediction.}
  \label{fig:transformer_results}
\end{figure}

We compare \systemname with a supervised transformer baseline to validate the zero-shot approach. The supervised model encodes each stay point as a feature vector that includes duration, start time, day of week, and POI activity types. It then processes sequences through a transformer encoder with MLP classification heads. Figure~\ref{fig:transformer_results} shows that the supervised baseline loses accuracy with less training data, while \systemname maintains competitive zero-shot performance and offers interpretable reasoning chains. These results demonstrate the practical benefits of LLM-based zero-shot inference when labeled data is limited or unavailable.

\small
\begin{table}[h]
\centering
\vspace{-.1in}
\caption{Experimental results for demographic inference on age, income, and education, evaluated using accuracy (Acc.) and F1 score (F1). Numbers in parentheses indicate the number of classification categories.}
\label{tab:main_results}
\begin{tabular}{l|cc|cc|cc}
\toprule
& \multicolumn{2}{c|}{\textbf{Age (4)}} & \multicolumn{2}{c|}{\textbf{Income (6)}} & \multicolumn{2}{c}{\textbf{Education (5)}} \\
\textbf{Model} & Acc. & F1 & Acc. & F1 & Acc. & F1 \\
\midrule
Mistral-7B + CoT & \textbf{0.442} & 0.318 & \textbf{0.293} & 0.133 & \textbf{0.254} & 0.123 \\
Qwen3-8B + CoT  & 0.366 & \textbf{0.330} & 0.285 & \textbf{0.143} & 0.238 & \textbf{0.189} \\
\bottomrule
\end{tabular}
\vspace{-.1in}
\end{table}
\normalsize

We conduct ablation experiments on income prediction with Mistral-7B to examine the contribution of each reasoning stage in the CoT pipeline. We select income because income has the largest number of categories, making it more challenging than age or education. The supervised baseline performs poorly on income, while the LLM-based method still achieves competitive performance. Table~\ref{tab:ablation_income} shows that removing either factual feature extraction or behavioral pattern analysis slightly reduces performance compared with the full CoT model. Both stages provide complementary information. The small drop shows the model retains partial reasoning ability when one stage is missing, while the full CoT model achieves the best performance.

\begin{table}[h]
\centering
\vspace{-0.1in}
\caption{Ablation study on Chain-of-Thought (CoT) reasoning components for income inference.
We compare the full three-stage CoT pipeline (Full-CoT) with variants that remove Stage 1 (No-S1) or Stage 2 (No-S2).
Results are evaluated using accuracy (Acc.) and macro F1 score (F1).}
\label{tab:ablation_income}
\vspace{-.1in}
\begin{tabular}{l|cc}
\toprule
& \multicolumn{2}{c}{\textbf{Income (6)}} \\
\textbf{Model Variant} & Acc. & F1 \\
\midrule
Full-CoT & \textbf{0.293} & \textbf{0.133} \\
No-S2                    & 0.290 & 0.131 \\
No-S1                   & 0.289 & 0.130 \\
\bottomrule
\end{tabular}
\vspace{-0.1in}
\end{table}
\normalsize

Future work includes validating \systemname on additional real-world datasets and LLMs to assess generalizability. We also plan to explore imbalance-aware strategies, such as reweighting or attention-shifting, to improve performance on minority classes. Finally, expanding baseline comparisons to include a broader range of state-of-the-art methods will help contextualize the practical advantages and limitations of zero-shot LLM-based demographic inference.

\vspace{-.1in}
\section*{Acknowledgments}
Supported by the Intelligence Advanced Research Projects Activity (IARPA) via the Department of Interior/Interior Business Center (DOI/IBC) contract number 140D0423C0033. The U.S. Government is authorized to reproduce and distribute reprints for Governmental purposes notwithstanding any copyright annotation thereon. Disclaimer: The views and conclusions contained herein are those of the authors and should not be interpreted as necessarily representing the official policies or endorsements, either expressed or implied, of IARPA, DOI/IBC, or the U.S. Government.

\vspace{-.1in}
\printbibliography
\end{document}